\documentclass[10pt,twocolumn,letterpaper]{article}

\usepackage{wacv}
\usepackage{times}
\usepackage{epsfig}
\usepackage{graphicx}
\usepackage{amsmath}
\usepackage{amssymb}
\usepackage{booktabs}
\usepackage{algorithm}
\usepackage[noend]{algpseudocode}
\usepackage{hyphenat}
\usepackage{float}

\usepackage{xcolor, soul}
\sethlcolor{green}

\algnewcommand{\Inputs}[1]{%
  \State \textbf{Inputs:}
  \Statex \hspace*{\algorithmicindent}\parbox[t]{.8\linewidth}{\raggedright #1}
}
\algnewcommand{\Initialize}[1]{%
  \State \textbf{Initialize:}
  \Statex \hspace*{\algorithmicindent}\parbox[t]{.8\linewidth}{\raggedright #1}
}

\def\wacvPaperID{2006} 

\wacvalgorithmstrack   

\wacvfinalcopy 

\ifwacvfinal
\usepackage[breaklinks=true,bookmarks=false]{hyperref}
\else
\usepackage[pagebackref=true,breaklinks=true,colorlinks,bookmarks=false]{hyperref}
\fi

\begin{document}

\title{Image Segmentation-based Unsupervised Multiple Objects Discovery}

\author{Sandra Kara
\and
Hejer Ammar
\and
Florian Chabot
\and
Quoc-Cuong Pham\\
\and
Universit\'e Paris-Saclay, CEA, List, F-91120, Palaiseau, France\\
{\tt\small \string{firstname.lastname\string}@cea.fr}
}

\maketitle
\thispagestyle{empty}

\begin{abstract}
Unsupervised object discovery aims to localize objects in images, while removing the dependence on annotations required by most deep learning-based methods. To address this problem, we propose a fully unsupervised, bottom-up approach, for multiple objects discovery. The proposed approach is a two-stage framework. First, instances of object parts are segmented by using the intra-image similarity between self-supervised local features. The second step merges and filters the object parts to form complete object instances. The latter is performed by two CNN models that capture semantic information on objects from the entire dataset. We demonstrate that the pseudo-labels generated by our method provide a better precision-recall trade-off than existing single and multiple objects discovery methods. In particular, we provide state-of-the-art results for both unsupervised class-agnostic object detection and unsupervised image segmentation.

\end{abstract}

\section{Introduction}

Deep learning methods have shown tremendous advances in resolving several computer vision tasks such as object detection and image segmentation. However, massive amounts of carefully labeled images are necessary to train reliable deep learning models that can reach high performances. Due to the high cost of such manual annotations, several approaches were proposed to use only limited amounts of annotated data, such as semi-supervised learning, weakly supervised learning or few-shot learning. In this work, we address the problem of localizing objects in images without any supervision, called Unsupervised Object Discovery (UOD). 

UOD can be useful for other vision tasks related to object localization. Pseudo-labels  generated without supervision have been shown to provide reliable object priors for image instance retrieval in \cite{DtD}. For object detection, they can be used either to initialize an object detector without additional annotation \cite{LOST}, or in a semi-supervised setting, when combined with few labeled data \cite{DDT+}. Providing robust pseudo-labels with limited noise is key to the success of these tasks. However, this remains a major challenge, especially in a completely unsupervised context, where no prior knowledge is provided about the semantics and localization of objects present in an image.

Many approaches solve this problem by leveraging inter-image similarities \cite{OSD, rOSD} between pairs of object proposals in different images. Without carefully designed optimization mechanisms, these methods come with a computational cost and complexity that compromise their scalability. Moreover, these methods have shown to be dependent on supervised CNN features for the calculation of similarities.

Recently, vision transformers (ViT) have achieved excellent performances, outperforming CNN architectures in both supervised tasks \cite{ret, ViTs, DETR} and self-supervised learning \cite{DINO, MOCOV3}. Particularly, strong objects localization hints emerge from training ViT models using a self-distillation scheme, in DINO \cite{DINO}. These self-supervised features were so far explored only for solving single object discovery task \cite{LOST,TokenCut}. TokenCut \cite{TokenCut} demonstrated the effectiveness of spectral-clustering applied on self-supervised ViT features, for saliency detection, and significantly improved the state-of-the-art for single object discovery.

In this work, we propose a new approach to address multiple objects discovery without any supervision. We explore self-supervised Vision Transformer (SS-ViT) features to localize and segment multiple object instances in the image. Discovering multiple objects in each image is not straightforward as it requires a clear definition of what an object is. In fact, objects are either defined as the annotated regions in the supervised setting, or as the salient region in each image, in unsupervised single object discovery approaches. To address the localization of multiple objects in a fully unsupervised way, we propose to recognise object regions using a semantic information captured at the dataset level. In other terms, an object is defined as belonging to one of the discovered semantic categories, in the image collection. Concretely, the semantic categories present in the dataset are discovered in an unsupervised way. This information is encoded using classification models. Object discovery is then designed as the activation of object parts in each image using SS-ViT features, and the merging of these object parts using the self-supervised classifiers, to discover complete object instances. The effectiveness of the proposed framework is demonstrated through extensive experiments on the object detection benchmarks PASCAL VOC \cite{VOC} and MSCOCO \cite{COCO}. Since by design, our method provides pixel-wise mask proposals, we also show that the same framework solves the unsupervised image segmentation task.

Our contributions can be formulated as follows :
\begin{itemize}
\item We propose a fully unsupervised, bottom-up approach, for multiple objects discovery. We first discover object parts using intra-image similarities. Object parts are merged using a dataset-driven information, to form complete object instances. Both stages exploit self-supervised ViT features to produce instance masks. To the best of our knowledge, this is the first work that builds on SS-ViT features to solve the multi-object discovery task.
\item We generalize the saliency-based approach in TokenCut \cite{TokenCut} for the discovery of local fine semantic concepts (object parts) of multiple objects in an image.
\item We propose a novel semantic object proposal method for the self-supervised learning of a region classifier. This visual model encodes the dataset-level semantic information.
\item We improve the state-of-the-art in unsupervised multiple objects discovery, unsupervised class-agnostic object detection and unsupervised image segmentation, on challenging object detection benchmarks.
\end{itemize}

\section{Related work}

\subsection{Unsupervised object discovery/co-localization} 
\label{sub:UOD}
We can distinguish, from previous works, two distinct tasks: object discovery and object co-localization. The former consists in localizing objects in an image without any prior knowledge of the image content. This is the \textit{real} object discovery task, which is much more challenging than object co-localization \cite{rOSD}. On the other hand, co-localization aims at localizing common objects between images, that share the same semantic content. Algorithms in this setting are fed with perfect image clusters derived from the ground-truth. It is therefore a weakly supervised version of object discovery.

DDT \cite{DDT+} addresses the co-localization task and is the first work that demonstrated the reusability of supervised CNN features for object co-localization. In DDT, objects are selected from regions of high correlation within a given cluster (semantic category). 

Other methods address both tasks, and many of them leverage inter-image similarities between off-the-shelf region proposals. Cho \textit{et al.} \cite{PM} formulated the problem as a structure and objects discovery, by iterating part-matching and object localization. Similarly, OSD \cite{OSD} simultaneously localizes objects and discovers structures of the image collection. It formalizes the task as an optimization problem. Although OSD brought a large improvement, it has shown to be highly dependent on supervised proposals provided by \cite{RP}. The method also suffers from overlapping region proposals, which prevents it from proposing multiple objects per image. These limitations were addressed in rOSD \cite{rOSD} by providing unsupervised proposals corresponding to regions of high activations around local maxima, within CNN feature maps. rOSD also constrains the number of proposals per local maximum, and performs non-maximum suppression (NMS) \cite{FasterRCNN} post-processing, to address the problem of overlapping proposals. Note that these methods, while unsupervised, are built on supervised CNN features, from the ImageNet \cite{imagenet} classification task. LOD \cite{LOD} formalized the task as a ranking problem, and focused on ensuring the scalability of the proposed approach. It also demonstrated the utility of self-supervised CNN features for single and multiple objects discovery.

Other methods \cite{LOST, TokenCut} tackled the single object discovery problem, and showed the potential of self-supervised features, especially from ViT models, for saliency detection. LOST \cite{LOST} proposed a seed expansion heuristic based on inter-patch correlation. TokenCut \cite{TokenCut} investigated the use of spectral-clustering on self-supervised ViT features, which are projected into a new space that allows for a more accurate binary clustering \cite{SC}. 

In previous multi-object discovery methods, relying on inter-image similarities in the computation of an objectness score could favor the discovery of the most frequent objects. In our approach, even though we also use a dataset-driven information, we overcome this issue by training visual classification models to encode the information of semantic classes. This results in a better separation between object and non-object regions, and a better detection of under-represented classes.

\subsection{Unsupervised image segmentation}

Image segmentation is the task of grouping all pixels of an image into meaningful regions, where pixels sharing the same characteristics are assigned to the same region \cite{DFC}. Due to the very high cost of such a dense annotation, weakly supervised and fully unsupervised methods were explored. In the weakly supervised setting, \cite{wekly_sup_segmnetation} takes as input the image-level labels of the class categories present in the image, and utilizes a vision-language embedding model to create a rough segmentation map for each class. 

Other approaches do not use any kind of supervision. On one hand, we find classic methods such as k-means \cite{kmeans}, that focuses on pixels clustering based on color and texture features, and assigns each pixel to the cluster with the nearest mean. Moreover, graph-based segmentation (GS) \cite{GS} generates image segments, while ensuring that these segments are not being too coarse or too detailed, based on regions comparison. More recently, methods based on unsupervised learning for image segmentation have been introduced. For example, IIC \cite{IIC} learns to maximize the mutual information between an image and its augmentations on a patch-level cluster. Kim \textit{et al.} \cite{superpixels, DFC} trains a CNN by iterating features clustering and network parameters tuning. The method is based on three criteria to maximize the features similarity between spatially continuous pixels and pixels assigned to the same cluster, while imposing a large number of clusters. The authors proposed two solutions for label assignment without any supervision (i) by superpixel extraction using simple linear iterative clustering in \cite{superpixels} and (ii) by the use of a spacial continuity loss in \cite{DFC} to address the limitation of fixed segment boundaries.

These methods discover multiple objects by proposing dense object candidates. Several other methods address the semantic segmentation task in an unsupervised way, without proposing dense object discovery. We do not consider these approaches in our study as we solve a different task.  

\subsection{Self-supervised vision transformers} 

The self-supervised setting aims at learning useful representations with no real label. It was first used for pre-training CNN models \cite{AE, CE, DAE}, and showed a strong generalization ability to downstream tasks. More recently, the self-attention based encoding of images using transformers for vision \cite{ViTs} was proved to be effective for a large spectrum of supervised vision tasks such as classification \cite{ViTs}, semantic segmentation \cite{Segmenter} and dense prediction tasks \cite{dense_prediction}. ViT has also become a reference architectural choice of neural nets for visual representation learning. In addition to the classic masked auto-encoding paradigm inspired from NLP, MoCo-v3 \cite{MOCOV3} among others, demonstrated a strong potential of training ViT with a contrastive approach. 

Recently, a self-distillation scheme was used in DINO \cite{DINO} to train ViT  with no labels. The choices made during training result in effective semantic separation and local-global alignment of the learned features. In particular, the resulting attention maps strongly activate the object regions, which provide clues to the localization of objects in the image. SS-ViT features have been explored recently to perform saliency detection and single object discovery tasks \cite{LOST, TokenCut}.

To the best of our knowledge, our method is the first to exploit self-supervised ViT features in a fully unsupervised multiple object discovery pipeline. The method outputs object instance masks resolving also the unsupervised image segmentation task. Such results can be used as pseudo-labels to initialize the training of a class-agnostic object detector, without any supervision.

\section{Method}

\subsection{Overview}

\begin{figure*}[h]
  \centerline{\includegraphics[width=17cm]{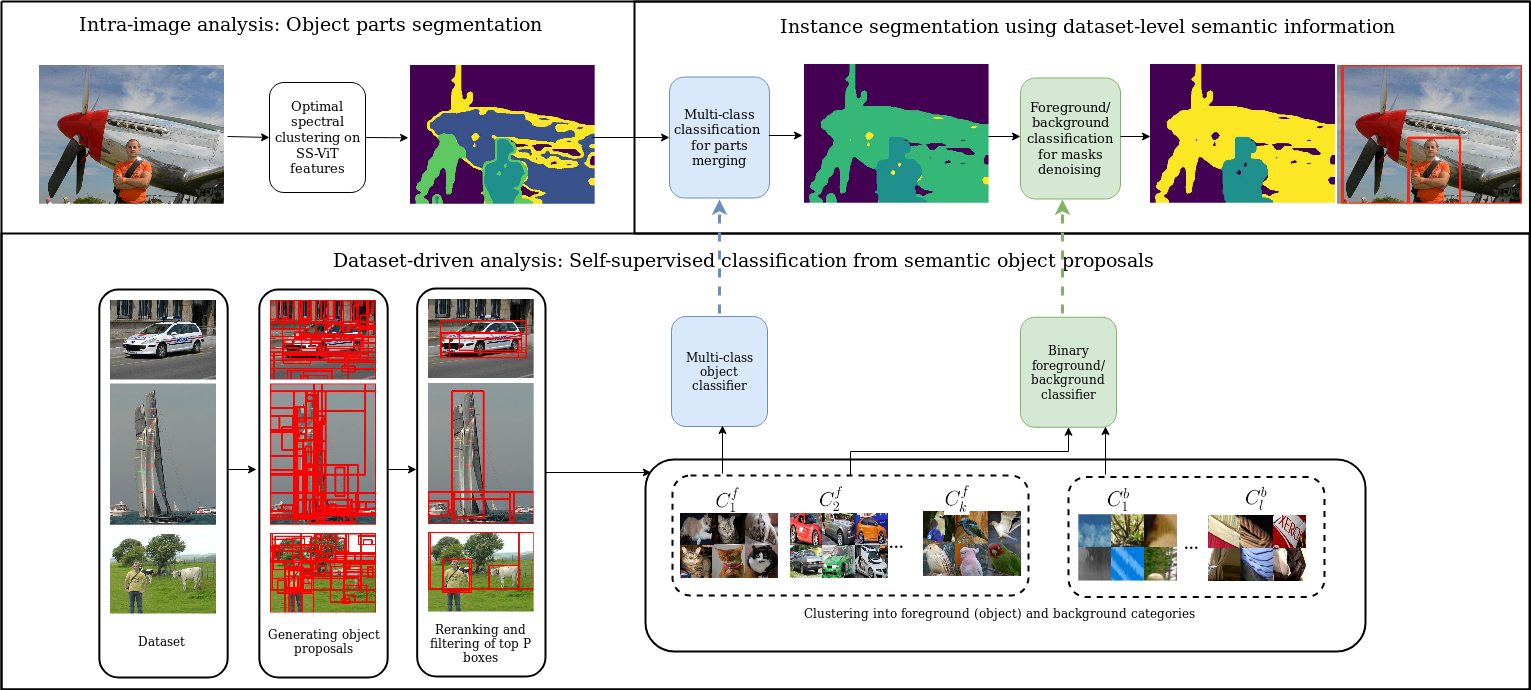}}
  \caption{\textbf{Pipeline of the method.} Top left: Intra-image analysis for the discovery of local semantic concepts. Bottom: Dataset-level analysis for the selection of semantic object proposals to train self-supervised classifiers. Top right: Using the data-driven classifiers on each image for parts merging and denoising.}
  \label{fig:global}
\end{figure*}

Recently, SS-ViT features showed to generalize well to saliency-based tasks \cite{LOST, TokenCut}. In this work, we aim to demonstrate the potential of using those features for multiple objects discovery, without any supervision. We adopt a bottom-up approach, illustrated in figure \ref{fig:global}, starting with an intra-image analysis, for the discovery of object parts. At the dataset level, two CNN models are trained in a self-supervised manner, using carefully selected, and semantic object proposals. These models are used to merge and filter object parts, to form complete object instances.

The intra-image analysis can be seen as a generalization of TokenCut \cite{TokenCut} to the multiple object discovery task. Similar to TokenCut, we perform spectral clustering using SS-ViT features, to decompose the image into eigen vectors with useful information. Different from TokenCut: (i) Since we focus on the localization of multiple objects, we look for more localization clues than just saliency. Thus we use multiple eigen vectors, as the feature space to apply local clustering, instead of only using the second eigen vector. (ii) The number of local clusters is no longer known as we try to solve the multi-object discovery task (2 clusters in saliency detection task). To manage this, we propose an algorithm for choosing an \textit{optimal} number of clusters, without any knowledge about the number of objects, or semantic concepts, in each image. The algorithm is detailed in section \ref{sub:local analysis} and aims at discovering multiple object parts, while limiting over-segmentation.

The goal of the dataset-level analysis is to build two classifiers that capture the dominant semantic classes in the image collection. One classifier is used for merging object parts, resulting from the local segmentation, and associates  a confidence score to each discovered object. The second classifier separates foreground/background classes and is used to filter remaining noise after the merging phase. We perform image clustering to get pseudo-labels for training both models. Since images may contain several semantic concepts, instead of using the whole images, we apply clustering on selected object proposals from Selective Search \cite{SeSe}. For proposals selection, we build an objectness score detailed in section  \ref{sub: object proposal}. The retained top proposals are grouped into clusters, which are used for training the classifiers.

Finally, the classifiers are used in cascade to merge and denoise the discovered object parts. Both stages use self-supervised ViT features trained using DINO \cite{DINO}. We show in section \ref{sub: object proposal} how these features are particularly relevant to our approach because of some properties like semantic separation, local-global alignment, and object regions activation.

\subsection{Discovery of intra-image semantic concepts} \label{sub:local analysis}

In this step, we extend TokenCut \cite{TokenCut} to discover multiple objects in each image, instead of solving  saliency detection. TokenCut constructs a weighted graph where nodes are ViT embeddings of image patches and edges correspond to the cosine similarity between tokens. Single object discovery is then formalized as a normalized graph-cut (Ncut) problem, which is solved using spectral clustering: features are projected into a new space via eigen decomposition. In this space, the second smallest eigen vector provides a solution to the Ncut problem for binary clustering, as demonstrated by Shi and Malik \cite{SC}. Likewise, we create a similarity graph based on SS-ViT features. The image is then decomposed into eigen vectors with useful information. We consider $N$ eigen vectors ($N \geq 2$) for local clustering, since we aim to capture multiple objects in the image. The choice of $N$ is studied in section \ref{sub: ablation study}. The selected $N$ eigen vectors represent the feature space where local clustering of image pixels is performed: each pixel is represented with a new feature vector $f'_i$ of size $N$, where $i$ varies between $1$ and the total number of pixels per image $n_p$.


Since in a fully unsupervised setting the number of semantic concepts in each image is unknown, we determine an optimal number of clusters $K$ using an iterative process as detailed in algorithm \ref{algorithm}. We apply k-means clustering to the image pixels, in the new feature space of eigen vectors $\mathcal{F} = \{f'_i; 1\leq i \leq n_p\}$. This  partitions the image into $K$ groups, which we denote $C_K$. We consider the background cluster as the one occupying the biggest area in the image. The background id is denoted as $b\_id$. All the remaining clusters represent the \textit{objects area}. $K$ is incremented, starting from $K=2$, until no significant \textit{object area} is newly activated. The goal is to activate multiple object regions in the image, while limiting over-segmentation. Examples of the results of this step are provided in figure \ref{fig:examples}, first column. In particular, we see in the last row that, in some cases, the algorithm directly outputs an optimal segmentation of the image. This shows its effectiveness compared to a simple over-segmentation, where a predefined number of clusters is used, without adapting to the content of each image.

\begin{algorithm}[h]
\caption{Iterative clustering for intra-image discovery of semantic concepts} \label{algorithm}

  \begin{algorithmic}[1]
  
  \Initialize{$K \gets 2$ \\
$C_K \gets K\-means(\mathcal{F}, K)$ \\
$b\_id \gets \underset{k}{\arg\max} \{area(C_K(k)), 1\leq k \leq K\} $ \\
$obj\_area \gets \sum_{k=1, k\ne b\_id}^{K} area(C_K(k))$ \\
$add\_semantic\_concepts \gets True$}

\While{$add\_semantic\_concepts$}
\State $K \gets K+1$
\State $C_K \gets K\-means(\mathcal{F}, K)$
\State $new\_obj\_area \gets \sum_{k=1, k\ne b\_id}^{K} area(C_K(k))$
\If{$ \dfrac{new\_obj\_area}{obj\_area}>thresh$}
    \State $obj\_area \gets new\_obj\_area$
\Else{}
    $ add\_semantic\_concepts \gets False$
\EndIf
\EndWhile
\State \Return $K$
\end{algorithmic}
\end{algorithm}

\subsection{Dataset-level semantic object proposals} \label{sub: object proposal}

As stated above, we use Selective Search \cite{SeSe} (SeSe) region proposals as object priors to discover the semantic classes in the dataset, through proposals clustering. These proposals provide a fairly high recall. However, their ranking is rather naive: given an over-segmentation of the image, the regions merged first, based on color and texture similarities, are ranked first. This makes even the top proposals subject to a lot of noise. We thus propose a new ranking of SeSe proposals, to select the most relevant ones. To do this, we build an objectness score, based on assumptions about object-like regions.

Note that the objectness score is computed within each image, independently from all other images in the dataset. Concretely, we use two main measures in this computation: intersection over union (IoU) and cosine similarity between object proposals in the same image. Given $M$ proposals ($p_{1}$, $p_{2}$,  ..., $p_{M}$) in an image, we define $u_{ij}$ as the overlap rate and $s_{ij}$ as the similarity between $p_{i}$ and $p_{j}$. For the latter, we use the cosine similarity between the $CLS$ tokens from the last layer of a ViT trained using DINO. Let $f_{i}$ and $f_{j}$ be the feature vectors ($CLS$ token) that result from passing $p_{i}$ and $p_{j}$ respectively to the SS-ViT. The cosine similarity $s_{ij}$ is defined as:
\begin{equation}
s_{ij} = \frac{f_{i}.f_{j}}{||f_{i}|| ||f_{j}||} 
\end{equation}

\noindent The new objectness score for object proposals re-ranking is the weighted sum of three normalized terms:

\begin{equation}
\resizebox{0.9\linewidth}{!}
{
$score(p_{i}) = \frac{\alpha}{2}(Sim_{L}(p_{i}) + Dissim_{G}(p_{i})) + (1-\alpha)H(p_{i})$}
\end{equation}
Each term of this score is based on a different assumption:

\noindent\textbf{Object-like regions have high local similarity.} We define local similarity for a given proposal $p_{i}$ as its average similarity to its neighbouring proposals, i.e. proposals having an IoU with $p_{i}$ above a threshold $t$. We notice that these proposals correspond usually to parts of the same objects. We also recall that we are using SS-ViT features learned using DINO, with a global local alignment objective. This means that an object is close to its parts in the DINO features space. From this we deduce that a high similarity between $p_{i}$ and its neighbours increases its chance of containing an object. Thus, we make all the neighbours of $p_{i}$ vote positively for it, in the following local similarity term:
\begin{equation}
Sim_{L}(p_{i}) = \sum_{j=1}^{M} s_{ij}  , j\neq i , u_{ij} \geq t
\end{equation}

\noindent\textbf{Object-like regions have high global dissimilarity.} We now consider the global similarity, i.e. the average similarity between $p_{i}$ and all other proposals that \textit{do not} overlap with $p_{i}$. Given the foreground/background imbalance in real-world images, most object proposals in an image occupy the background, and have similar visual content (e.g. 'sky'). Objects, on the contrary, occupy regions that are distinct in the image. If a proposal $p_{i}$ contains an object, then it has low overall similarity, as all object proposals in the background vote negatively for it. Thus, $p_{i}$ will be highly dissimilar, hence the following global dissimilarity term:
\begin{equation}
Dissim_{G}(p_{i}) = \sum_{j=1}^{M} (1-s_{ij})  , j\neq i , u_{ij}<t
\end{equation}

\noindent\textbf{Object-like regions have high entropy.} The Shannon entropy of a discrete random variable is defined as:
\begin{equation}
H(p) = - \sum_{x} P_{x}log(P_{x}) 
\end{equation}
This measure is used to quantify the randomness of a variable \cite{shannon}. In image processing, $P_{x}$ refers to the distribution of gray levels $x$ in image $p$ (or colors intensities in RGB images). The previous formula associates higher entropy to images with more details and colors variation. Inversely, homogeneous regions are characterized by a low entropy. We thus associate low-entropy proposals to background, by adding an entropy term in the final objectness score. 

\begin{figure*}[h]
  \centerline{\includegraphics[width=16cm]{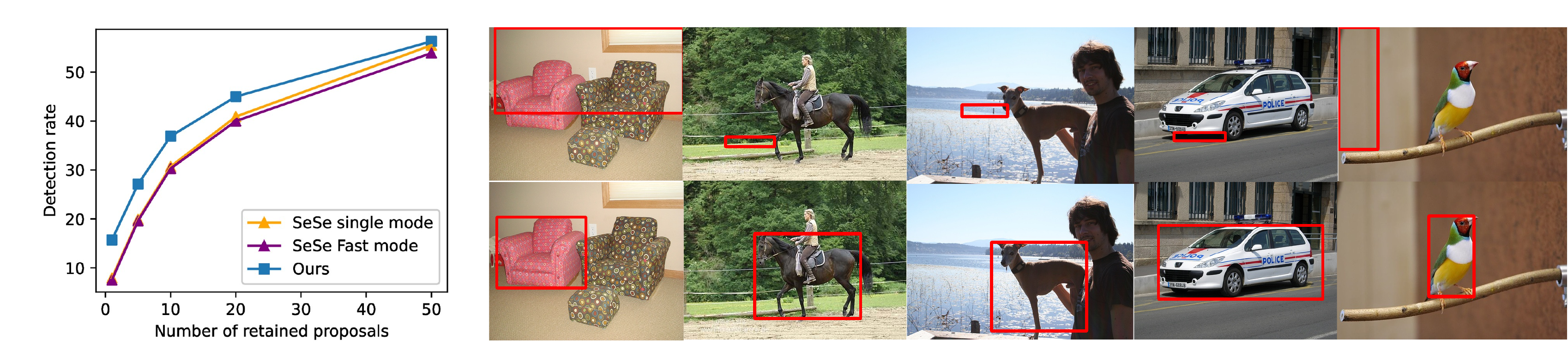}}
  \caption{\textbf{Results of ranking object proposals.} Left: Comparison of the detection rate given the number of retained top proposals from two modes of SeSe, with our ranking. Right: Examples of the top1 proposal using SeSe score (top) and our score (bottom).}
  \label{fig:score}
\end{figure*}

It can be seen from figure \ref{fig:score} that the proposed ranking improves the detection rate for a fixed number of proposals, compared to two modes of SeSe. This is especially true when a small number of proposals are selected. We also compare qualitatively the retained top\hyp{}1 proposals with the two rankings.

We recall that the aim of this new ranking is to reduce the amount of noise in the top proposals, which will be retained for clustering, as explained in section \ref{sub: global clustering}. We choose to use SeSe proposals for its popularity. However, the proposed ranking should be valid with other proposals, provided that they have a similar distribution, i.e. bounding boxes that occupy the whole image, and thus verify the background dominance condition, discussed above.

\subsection{Dataset-driven self-supervised region classification} \label{sub: global clustering}

After re-ranking the SeSe object proposals, the goal is to transform the top P object priors in each image into pseudo-labels to train multi-class classifiers. These will be used to merge and refine the discovered local semantic concepts. We use a value of P large enough to make sure that we do not only select objects but also object parts. This is important since the classifier must learn to assign the same semantic class to an object and its parts, for an accurate merging.  We use k-means clustering \cite{kmeans} on the SS-ViT features of all the selected object proposals. The optimal number of clusters is chosen by finding the best silhouette score \cite{silhouetteCluster}, which minimizes the mean intra-cluster distance and maximizes the mean nearest-cluster distance. With the semantic information contained in SS-ViT features, similar semantic concepts are grouped together. Moreover, each cluster contains proposals of both objects and their parts. This is especially due to the multi-crop augmentation technique used in DINO. The obtained clusters capture the global semantic information of the dataset. Note that the number of the clusters is not necessarily equal to the number of classes annotated in the dataset. However, we can still localize instances of undiscovered categories, such as `bottle' and `plant'.\newline

Since some of the selected proposals may still belong to background (Bg) regions, some of the discovered pseudo-classes are Bg clusters, that we aim to identify. According to our ranking score detailed in \ref{sub: object proposal}, the proposals having the lowest scores are the ones representing most probably Bg regions. 
Each of these proposals is passed to a SS-ViT to extract its features. The average vector of these features is considered as a \textit{pattern} of Bg regions. 
The clusters whose center has a distance below a threshold $t_{bg}$ with the Bg \textit{pattern}, are considered Bg clusters.\newline 

After Foreground (Fg) and Bg groups identification, we associate to the clusters two types of labels (i) Each Fg cluster is assigned an \textit{id} representing one discovered semantic class. (ii) All clusters have a binary label indicating whether it belongs to Fg or Bg. These image clusters are used to train two CNN-based classifiers, with the cluster \textit{id} as a classification target.
The first is a multi-class classifier trained using Fg clusters to assign objects and object parts to a specific class. The second classifier is trained using all the discovered clusters, and learns to distinguish between objects and Bg regions.

\subsection{Instance segmentation using dataset-level information}

\begin{figure}[h]
\begin{center}
  \centerline{\includegraphics[width=1\linewidth]{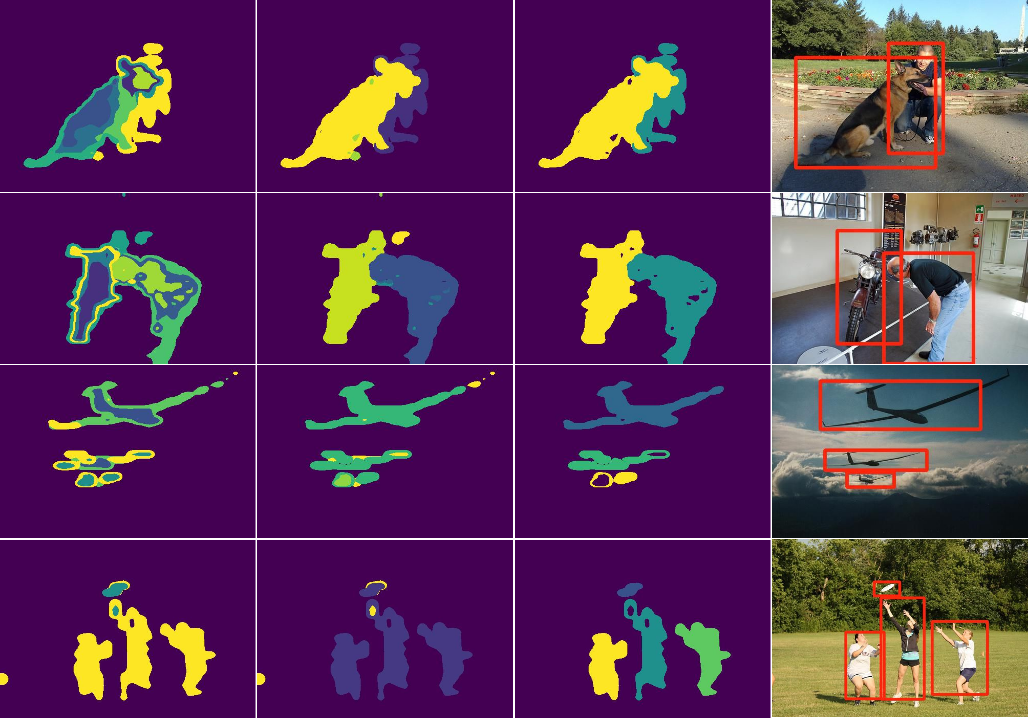}}
\end{center}
   \caption{\textbf{Example of results.} By column: results of the discovery of local concepts, segmentation result after parts merging, final instance mask segmentation, final bounding boxes.}
\label{fig:examples}

\end{figure}

In this final step, the obtained classifiers are used to merge and refine the object parts identified in the intra-image analysis. The multi-class classifier is first used on each segmented region: Image crops enclosing each object part segment are passed to the CNN-based classifier. Nearby regions assigned to the same category are merged to form complete object instances. The image crop around each merged region segment is then passed to the Fg/Bg classifier to eliminate segments classified as Bg. This binary classification is performed second to avoid incorrect classification of small object parts as Bg, if used before merging. The multi-class classifier also assigns to each object a confidence score, which is necessary for the evaluation metrics (AP@50, odAP). We provide in figure \ref{fig:examples} illustrations for each step of the proposed framework.

\section{Experiments}
\subsection{Implementation details}
Following previous works \cite{rOSD,LOD,LOST} we conduct our experiments on three detection and localization benchmarks: VOC2007 trainval, VOC2012 trainval \cite{VOC} and COCO20k which is composed of 19817 images randomly chosen from COCO2014 trainval dataset \cite{LOST}. We specify in the following the implementation details and hyper-parameters for each addressed task.

\noindent\textbf{Unsupervised multiple objects discovery}. In the intra-image analysis, local clustering is applied on SS-ViT features learned using the DINO training scheme. Based on the conclusions from previous works \cite{TokenCut,LOST}, we use the variant ViT-S with a patch size of $16$. Eigen decomposition is performed using the $keys$ features of the last layer. To find the optimal number of local clusters, we set as convergence criterion a fraction of newly activated area of $2\%$, ($thresh=1.02$ in algorithm \ref{algorithm}). The number of eigen vectors used for clustering is studied in section \ref{sub: ablation study}, and showed to be invariant to the dataset: 3 eigen vectors for PASCAL VOC and COCO20k. For object proposals re-ranking, the three terms are found to have an equivalent impact on the final re-ranking, with $\alpha=0.7$, and $t=0.1$. Proposals from Selective Search single mode are used in this work. For the dataset-level analysis, $P=20$ top proposals in each image are selected to train the classifiers. A distance threshold $t_{bg}=0.8$ from the Bg is used to separate Fg and Bg clusters. We use ResNet50 as the backbone of the two classifiers, initialized with DINO pre-training.

\noindent\textbf{Unsupervised class-agnostic object detection.} We follow the same configuration described in \cite{LOST} for training a class-agnostic Faster-RCNN, with our pseudo-labels. We also use the same batch-size and the number of training iterations, for an objective comparison with previous works.

\noindent\textbf{Unsupervised image segmentation.} Following \cite{DFC}, this experiment is conducted on VOC2012 validation set, consisting of 1446 images. Masks resulting from multi-object discovery task are evaluated using mIOU, see section \ref{sub: metrics}.

\subsection{Metrics and evaluation settings} \label{sub: metrics}

Different metrics are used to evaluate different tasks:

\noindent\textbf{Unsupervised multiple objects discovery.} Most of multiple objects discovery methods are based on ranking of object proposals. This makes them able to produce a large number of object candidates. The question then arises as to how many proposals to keep for computing recall, precision, or even the classical AP50 metric, since all of these would be affected by the number of retained top proposals. \cite{LOD} addressed this issue and proposed an new version of AP, adapted to the object discovery task, called odAP. odAP is presented as the area under the precision-recall curve, where each precision-recall point is computed for a number of retained proposals, starting from 1, to the maximum number of objects in any image in the dataset. Even though by design, our approach outputs a reduced number of proposals, we use odAP to compare with previous works. We report the odAP50 where a detection is considered correct if its overlap rate with a ground truth bounding box is above $50\%$. And odAP@$[50:95]$, which is the average odAP for 10 values of IoU, varying from $50\%$ to $95\%$.

\noindent\textbf{Class-agnostic unsupervised object detection.} A classical class agnostic Average Precision (AP@50) is calculated.

\noindent\textbf{Unsupervised image segmentation.} Following \cite{DFC}, we use the mean intersection over union (mIoU) to evaluate unsupervised image segmentation. mIOU is calculated as the average IOU between each ground truth mask (along with the background) and the detected mask that has the largest IOU with it, without considering any class label. 

\subsection{Unsupervised multiple objects discovery} \label{sub: multiple_discovery}

We follow previous works and evaluate our method using odAP  \ref{sub: metrics}. Note that this metric is particularly adapted to the methods that propose a large number of object candidates, based on a ranking of object proposals. Since our approach is built on image segmentation, a limited number of boxes are proposed: 3 per image on average in PASCAL VOC dataset \cite{VOC}. Hence, our approach is disadvantaged by this metric regarding the recall. Despite that, we show in table \ref{odAP} the superiority of our method on both odAP@50 and the much more demanding odAP[50-95] metric.

The higher odAP[50-95] demonstrates  the  accuracy  of  our  returned pseudo-boxes: Since these are generated from instance masks, they better enclose objects, and thus remain valid for a higher IoU threshold condition. Also, our method uses self-supervised features, which makes it  fully  unsupervised,  unlike  previous methods, which showed to be dependant on supervised features.

\begin{table}[h] 
  \begin{center}
    {\scalebox{0.58}{
    \begin{tabular}{lccccccc}
\toprule
\multicolumn{1}{l}{Method} &
\multicolumn{1}{l}{Features}&
\multicolumn{3}{c}{odAP@50}    & \multicolumn{3}{c}{odAP@{[}50-95{]}} \\
 \midrule
\multicolumn{1}{c}{} &
\multicolumn{1}{c}{} & \multicolumn{1}{c}{VOC07} & \multicolumn{1}{c}{VOC12} & \multicolumn{1}{c}{COCO20k} & \multicolumn{1}{c}{VOC07} & \multicolumn{1}{c}{VOC12} & \multicolumn{1}{c}{COCO20k} \\
\midrule
\emph{Kim et al.} \cite{kim2009unsupervised, LOST} & Sup & 9.5  & 11.8 & 3.93 & 2.5 & 3.1 & 0.96 \\
DDT+ \cite{DDT+, LOST} & Sup  & 8.7  & 11.1 & 2.41 & 3.0 & 4.1 & 0.73 \\
rOSD \cite{rOSD, LOST} & Sup   & 13.1 & 15.4 & 5.18 & 4.3 & 5.3 & 1.62 \\
LOD  \cite{LOD, LOST} & Sup            & 13.9 & 16.1 & \textbf{6.63} & 4.5 & 5.3 & 1.98 \\
Ours              & Self & \textbf{15.4} & \textbf{17.6} & 5.44 & \textbf{6.8} & \textbf{8.1} & \textbf{2.11} \\
\bottomrule
\end{tabular}
}}
\end{center}
\caption{Multi-object discovery performance in odAP (Average Precision for object discovery) }
 \label{odAP}
\end{table}

\subsection{Class-agnostic unsupervised object detection} \label{sub: object_detection}

State-of-the-art multiple objects discovery methods (MOD) usually rely on a ranking of object proposals based on inter-image similarities. These methods output a large number of object candidates and the question then arises as to how many bounding boxes to keep for the initialization of an object detector. Inversely, single object discovery methods (SOD) have a very limited recall. We argue that our method provides a better precision/recall trade-off than the previous methods in both settings. To prove this, we train a class-agnostic object detector using our generated pseudo-labels. Results are presented in table \ref{CAD}.
We notice a clear improvement with our approach compared to MOD methods on all tested datasets. The gap however gets smaller when comparing with the SOD methods on PASCAL VOC \cite{VOC} dataset. This can be explained by the presence of a dataset bias in PASCAL VOC: A large proportion of images in this dataset contain one object, which gives a clear advantage to SOD methods.
On the more challenging COCO20k dataset, our method exceeds both categories (MOD and SOD). This demonstrates the superiority of our pseudo-labels, even for datasets with complex scenes.

\begin{table}[h]
  \begin{center}
    {\scalebox{0.7}{
\begin{tabular}{lccc}
\toprule
Method           & VOC07 & VOC12 & COCO20K \\
\midrule
Selective Search \cite{SeSe} & 3.6 & 4.8 & 1.8 \\
EdgeBoxes \cite{EB}       & 2.9 & 4.2 & 1.6 \\
\midrule
rOSD + CAD  \cite{rOSD}      & 24.2 & 29.0 & 8.4 \\
LOD   + CAD \cite{LOD}     & 22.7 & 28.4 & 8.8 \\
\midrule
LOST   + CAD  \cite{LOST}   & \textbf{29.0} & 33.5 & 9.9 \\
TokenCut + CAD  \cite{TokenCut}   & 26.2 & 35.0 & 10.5 \\
\midrule
Ours  + CAD      & 27.9 & \textbf{36.2} & \textbf{13.8}\\
\bottomrule
\end{tabular}
}}
\end{center}
\caption{Class-agnostic unsupervised object detection in AP50\%} \label{CAD}
\end{table}

\subsection{Unsupervised image segmentation} \label{sub: image segmentation}

We further evaluate the performance of our method on VOC12 \cite{VOC} validation set for unsupervised image segmentation task (see Table \ref{segmentation}). Our method significantly outperforms previous state-of-the-art methods for discovering object masks in a fully unsupervised way. More qualitative results are provided in the supplementary material.

\begin{table}[h]
  \begin{center}
    {\scalebox{0.7}{
\begin{tabular}{ll}
\toprule
Method & VOC12 \\
\midrule
k-means clustering \cite{kmeans}, k=2 & 0.3166  \\
k-means clustering \cite{kmeans}, k=17 & 0.2383 \\
Graph-based segmentation (GS) \cite{GS}, $\tau$ = 100 & 0.2682 \\
Graph-based segmentation (GS) \cite{GS}, $\tau$ = 500 & 0.3647 \\
IIC \cite{IIC}, k=2 & 0.2729\\
IIC \cite{IIC}, k=20 & 0.2005\\
Kim \textit{et al.} with superpixels \cite{superpixels}  & 0.3082\\
Kim \textit{et al.} with continuity loss \cite{DFC}, $\nu$ = 5 & 0.3520\\
Ours &  \textbf{0.4247}\\
\bottomrule
\end{tabular}}}

\end{center}
\caption{Unsupervised image segmentation results in mIOU} \label{segmentation}
\end{table}

\subsection{Ablation study} \label{sub: ablation study}

In table \ref{score_ablation}, we provide an ablation study on different terms of the ranking score presented in \ref{sub: object proposal}. We evaluate the recall@50 (recall at IoU=50\%) for different numbers of the retained top proposals. We compare the results of the overall ranking score, with the ranking obtained when one of the terms is removed from the final score. The best results are achieved by considering all 3 terms, which supports the assumptions made in section  \ref{sub: object proposal}. We also compare our score with the original SeSe ranking of proposals from two settings. Using this new ranking, we ensure that the top proposals are more reliable for the classifiers training.

\begin{table}[h] 
  \begin{center}
    {\scalebox{0.7}{
    \begin{tabular}{llllllll}
\toprule
\multicolumn{1}{c}{{Method}} &  \multicolumn{4}{c}{Recall@50}    & \\
 \midrule
\multicolumn{1}{c}{{Number of boxes}} &  \multicolumn{1}{c}{1} & \multicolumn{1}{c}{4} & \multicolumn{1}{c}{10} & \multicolumn{1}{c}{20} \\
\midrule
SeSe Fast mode  \cite{SeSe}  &  7.5  & 19.6 & 30.3 & 40.0 \\
SeSe Single mode \cite{SeSe}  &  7.9  & 19.9 & 30.7 & 40.9 \\
\midrule
Ours: $Sim_{L} + Dissim_{G}$    &  13.9 & 23.5 & 34.4 & 44.1 \\
Ours: $Sim_{L} + H$          &  12.8 & 24.9 & 35.9 & 44.6 \\
Ours: Overall score     &  \textbf{15.7} & \textbf{27.1} & \textbf{36.9} & \textbf{45.0} \\
\bottomrule
\end{tabular}
}}
\end{center}
\caption{Ablation study on the impact of the different terms composing the re-ranking score, evaluated on VOC07 testset}
 \label{score_ablation}
\end{table}

We also provide a study of the number of eigen vectors to be used in the intra-image analysis, in order to activate multiple objects, while limiting the amount of noise. In table \ref{ablation_eigen}, we evaluate the AP@50 of the generated pseudo-boxes, to choose the best precision/recal trade-off. We conduct the study on PASCAL VOC and COCO since they present different distributions of objects. Considering this study, the reported results are obtained with 3 eigen vectors in the intra-image analysis, for both datasets.

\begin{table}[h]
  \begin{center}
    {\scalebox{0.7}{
\begin{tabular}{ccc}
\toprule
Number of eigen vectors & VOC07  & COCO20k\\
\midrule
2 & 22.1 & 5.9  \\
3 & \textbf{22.5} & \textbf{6.3}  \\
4 & 21.3 & 6.0  \\
5 & 20.3 & 5.8  \\

\bottomrule
\end{tabular}
}}
\end{center}
\caption{AP@50 as a function of the number of eigen vectors used for local analysis} \label{ablation_eigen}
\end{table}

\section{Conclusion and future work}
We presented a fully unsupervised approach for multiple objects discovery. The aim of this work was to address some of the limitations observed in existing methods. Namely, low recall in saliency detection-oriented methods, and the high amount of noise when several object candidates are proposed. We have shown that formulating the problem as an unsupervised segmentation is particularly suitable for reducing the noise in the generated pseudo-boxes. This provides a better precision-recall trade-off, which leads to a better initialization of an object detector. Still in this direction, we can further investigate the use of these pseudo-labels as an initial seed in a pseudo-labelling approach. Similarly, we can investigate the use of these object candidates with noise handling mechanisms.

\vfill
\section{Acknowledgements}
This work benefited from the FactoryIA supercomputer financially supported by the Ile-deFrance Regional Council.

{\small
\bibliographystyle{ieee_fullname}
\bibliography{egbib}
}

\clearpage

\onecolumn
\appendix
\section*{Supplementary material}
\addcontentsline{toc}{section}{Appendices}
\renewcommand{\thesubsection}{\Alph{subsection}}

\subsection{Evaluation of single object discovery methods on the multiple objects discovery task}

In our study, we have compared our approach mainly to the multi-object discovery (MOD) methods. The comparison with single object discovery (SOD) approaches has only been performed after training an object detector with pseudo labels from each method. Here we raise the question of what happens if SOD models are applied to MOD tasks. To answer this question in a quantitative aspect, we provide in table \ref{MOD} a comparison of our method with LOST\cite{LOST} and the more recent baseline TokenCut\cite{TokenCut}. Since these methods output a single object, odAP is no longer adapted and we use the AP50 metric for comparison. Note that AP50 is more relevant when a fixed number of objects are returned. Considering these results (table \ref{MOD}) and those of table \ref{odAP}, we conclude that our approach provides a better precision-recall trade-off than existing object discovery methods.

\begin{table}[H]
  \begin{center}
    {\scalebox{0.8}{
\begin{tabular}{llll}
\toprule
Method           & VOC07 & VOC12 & COCO20k \\

\midrule
LOST     [24]   & 15.7 & 17.7 & 4.1 \\
TokenCut    [31]   & 18.6 & 22.6 & 5.8 \\
Ours       & \textbf{22.5} & \textbf{23.1} & \textbf{6.3}\\
\bottomrule
\end{tabular}
}}
\end{center}
\caption{Multiple objects discovery results in AP@50\%} \label{MOD}
\end{table}

\begin{figure}[h]
  \centerline{\includegraphics[width=7.5cm]{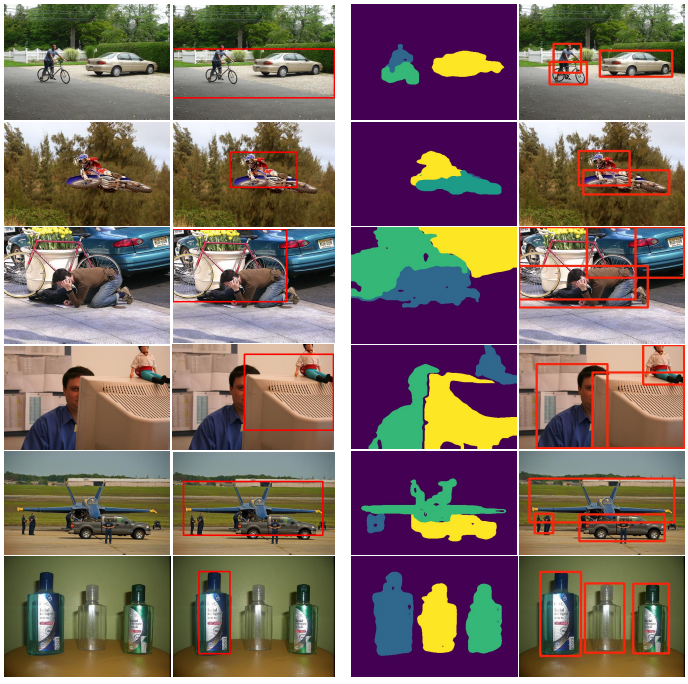}}
  \caption{\textbf{Qualitative results on VOC12 of the multiple objects discovery.} By column: original image, the predicted bounding-box from LOST\cite{LOST}, our segmentation result, our pseudo-boxes.}
  \label{fig:res1}
\end{figure}

\begin{figure}[H]
  \centerline{\includegraphics[width=7.5cm]{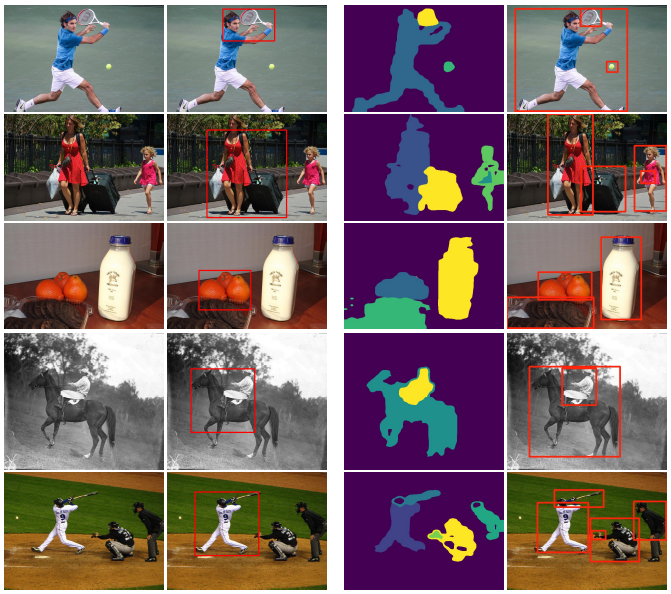}}
  \caption{\textbf{Qualitative results on COCO20K of the multiple objects discovery.} By column: original image, the predicted bounding-box from LOST\cite{LOST}, our segmentation result, our pseudo-boxes.}
  \label{fig:res2}
\end{figure}

\begin{figure}[H]
  \centerline{\includegraphics[width=12cm]{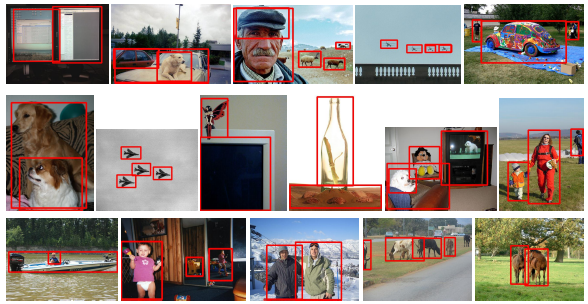}}
  \caption{Qualitative results for class-agnostic object detection on VOC12 dataset.}
  \label{fig:res3}
\end{figure}

\begin{figure}[H]
  \centerline{\includegraphics[width=12cm]{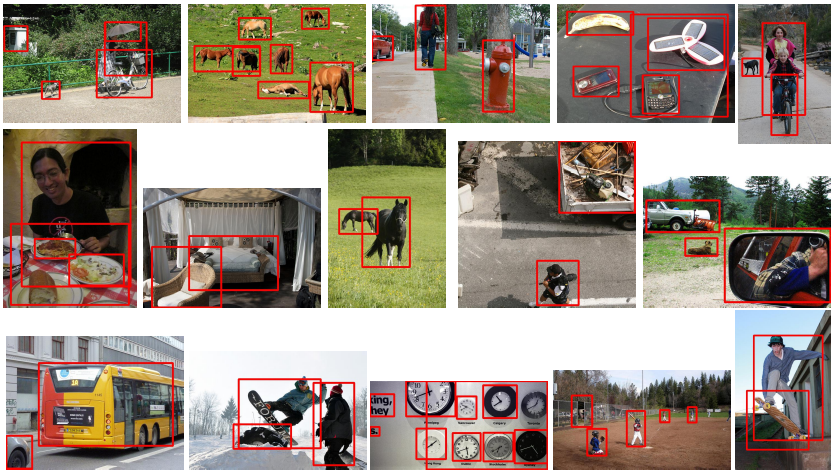}}
  \caption{Qualitative results for class-agnostic object detection on COCO20k dataset}
  \label{fig:res4}
\end{figure}

In the following, we provide additional qualitative results on VOC12 and COCOC20k datasets, for the tasks addressed in this work. In particular, in figures \ref{fig:res1}, \ref{fig:res2}, we compare our results with LOST\cite{LOST}, a single object discovery method that achieves the second best mAP after training the class-agnostic object detector (see table \ref{CAD}). We can see from the visual results that LOST usually groups several instances, even of different semantics, in complex scenes, with spatially close objects. This explains the superiority of our approach in the initialization of the class-agnostic object detector.  
We also show in figures \ref{fig:res3}, \ref{fig:res4} that after training an object detector with our pseudo-labels, we improve the separation of instances, even of the same semantic category.

\subsection{More qualitative results for multiple objects discovery on VOC12 dataset}

\subsection{More qualitative results for multiple objects discovery on COCO20k dataset}

\subsection{Qualitative results for class-agnostic object detection on VOC12 dataset}

\subsection{Qualitative results for class-agnostic object detection on COCO20k dataset}

\subsection{Failure cases}
\begin{figure}[H]
  \centerline{\includegraphics[width=9.5cm]{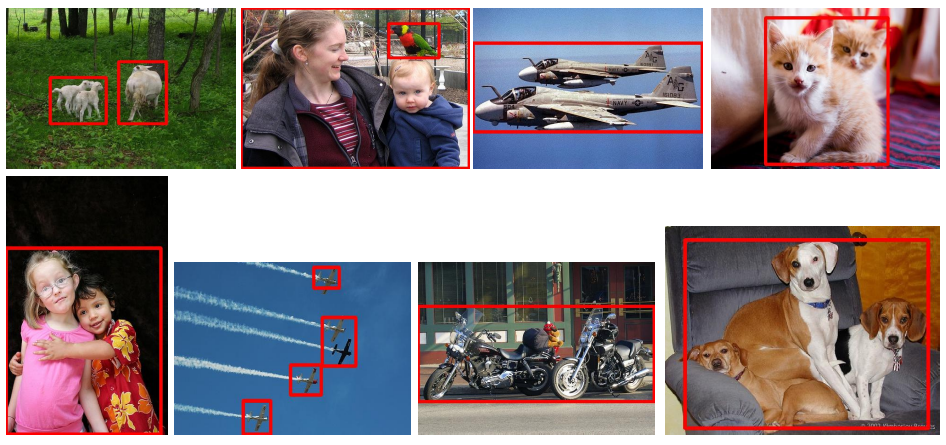}}
  \caption{Examples from VOC07 dataset where our approch fails to discover objects, by merging nearby instances of the same category.}
  \label{fig:fc}
\end{figure}

Our approach, although effective, shows some limitations that can be investigated in a future work. In particular, in the merging process, since nearby segments belonging to the same class are merged, this leads to the merging of nearby instances of the same category, see figure \ref{fig:fc}. This problem occurs in all existing object discovery methods \cite{rOSD, LOST, TokenCut}, and comes from the semantic information contained in both supervised and self-supervised features. Solving this problem requires, for example, the learning of an instance-variant representation, which is a challenging task. We hope that this work will stimulate interest in this research direction.

\end{document}